\documentclass[11pt,a4paper]{article}
\usepackage[hyperref]{acl2021}
\pdfoutput=1
\usepackage{times}
\usepackage{verbatim}
\usepackage{amsmath}
\usepackage{verbatim}

\usepackage{graphicx}
\usepackage{latexsym}

\usepackage{microtype}
\usepackage{multirow}

\aclfinalcopy % Uncomment this line for the final submission
\title{Check It Again: Progressive Visual Question Answering \\via Visual Entailment}
\author{Qingyi Si$^{1,2}$,\ Zheng Lin$^1$\thanks{\ \ Corresponding author: Zheng Lin.}\ ,\ Mingyu Zheng$^{1,2}$,\ Peng Fu$^1$,\ Weiping Wang$^1$ \\

$^1$Institute of Information Engineering, Chinese Academy of Sciences, Beijing, China \\
$^2$School of Cyber Security, University of Chinese Academy of Sciences, Beijing, China \\
  \texttt{\{siqingyi,linzheng,zhengmingyu,fupeng,wangweiping\}@iie.ac.cn} \\
   }
\date{}

\begin{document}
\maketitle
\begin{abstract}

While sophisticated Visual Question Answering models have achieved remarkable success, they tend to answer questions only according to superficial correlations between question and answer. %, without understanding the visual contents. 
Several recent approaches have been developed to address this language priors problem. However, %they fail to predict the correct answer according to one best output %they have not thoroughly crack the bias problem, and there is still room for improvement: 
%and there is still room for improvement:
%1) The prediction space is too large, including huge number of answers unrelated to the question. 2) 
most of them predict the correct answer according to one best output without checking the authenticity of answers. Besides,
they only explore the interaction between image and question, ignoring the semantics of candidate answers.
%The semantics of candidate answers have not been made good use of. 
%These enlighten us to shrink the prediction space and then check the authenticity of selected candidate answers with the help of their semantics. 
In this paper, we propose a select-and-rerank (SAR) progressive framework based on Visual Entailment. Specifically, we first select the candidate answers relevant to the question or the image, %-related candidate answers to shrink the prediction space, 
then we rerank the candidate answers by a visual entailment task, which verifies whether the image semantically entails the synthetic statement of the question and each candidate answer. Experimental results show the effectiveness of our proposed framework, which establishes a new state-of- the-art accuracy on VQA-CP v2 with a $7.55\%$ improvement.\footnote{The code is available at \url{https://github.com/PhoebusSi/SAR}}

\end{abstract}

\section{Introduction}
Visual Question Answering (VQA) task is a multi-modal problem which requires the comprehensive understanding of both visual and textual information. Presented with an input image and a question, the VQA system tries to determine the correct answer in the large prediction space. %This task is a challenging task which lies at the intersection of Computer Vision (CV) and Natural Language Processing (NLP), VQA pushes the boundary of both research fields. % and measures the extent to which computers have similar capabilities to humans.
Recently, some studies \citep{jabri2016revisiting,agrawal2016analyzing,zhang2016yin,goyal2017making} demonstrate that VQA systems suffer from the superficial correlation bias (i.e. language priors) caused by accidental correlations between answers and questions. As a result, traditional VQA models always output the most common answer\citep{selvaraju2019taking} of the input sample's question category, no matter what image is given. To address this language priors problem, various approaches have been developed.
%For example, VQA models could answer almost half of questions in the validation set without seeing the image\citep{manjunatha2019explicit}. This artificially high performance is problematic, and accordingly, various approaches have been developed to address the bias problem. 

 However, through exploring the characteristics of the existing methods, we find that whether the general VQA models such as UpDn\citep{anderson2018bottom} and LXMERT\citep{tan2019lxmert} or models carefully designed for language priors, as LMH\citep{clark2019don} and SSL\citep{zhu2020overcoming}, yield a non-negligible problem.  %Both models fail to predict the correct answer according to one best output. %have not thoroughly crack the bias problem. 
Both models predict the correct answer according to one best output without checking the authenticity of answers.
%There is still room for improvement: 
%On the one hand, the prediction space is large, including thousands of categories. On the other hand, 
Besides, these models have not made good use of the semantics information of answers that could be helpful for alleviating the language-priors.%the bias problem has not been thoroughly cracked, and there is still room for improvement. 

%%%%%%%%%%%%%%%%%%%%%%%%%%%%%%%%%%%%%%%%%%%%%%%%%%%%%%
\begin{figure}
\resizebox{0.95\linewidth}{!}{
  \centering
  \includegraphics[width=1.0 \linewidth]{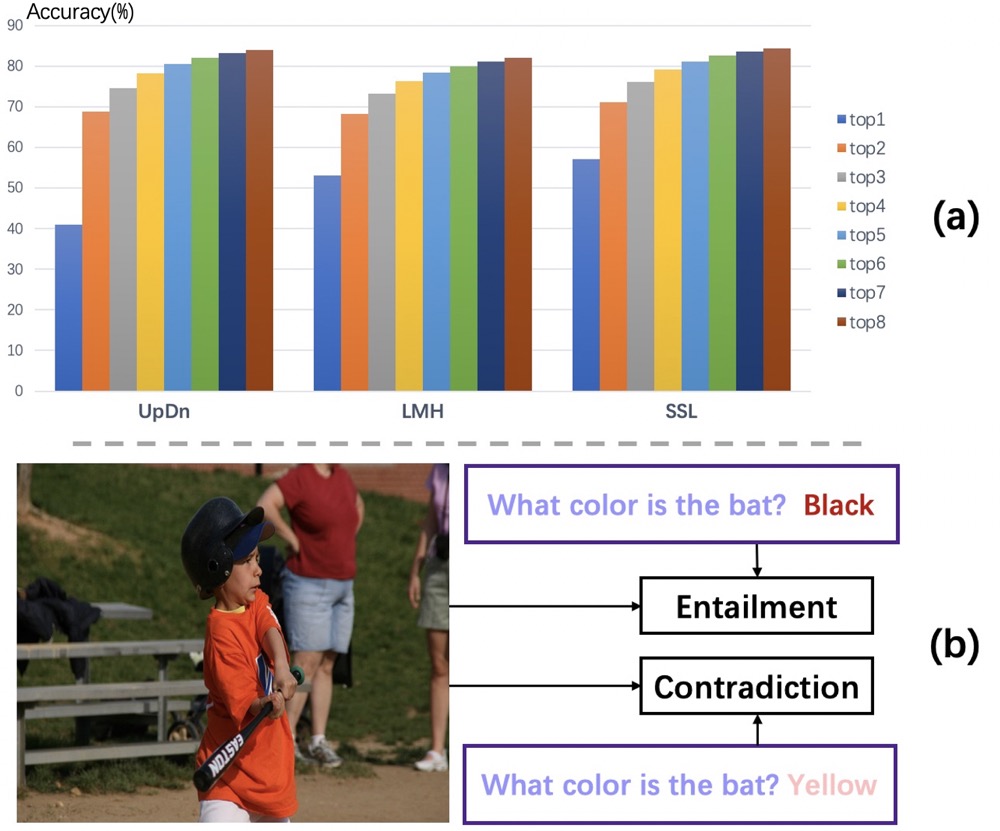}
  }
\caption{(a) We evaluate the performance of UpDn, LMH, SSL on the VQA-CP v2 test. $topN$ represents the topN accuracy.
%In the VQA-CP v2 test set, we divide correct answers into groups by their rankings of prediction probability, and $N$ represents correct answers which rank $N$-th when the model makes prediction
%we sort all candidate answers of each question by prediction probability, and N represents the correct answers which rank N-th. 
%(e.g. the orange cubes represent the difference between top2 acc and top1 acc). %proportion of the correct answer having second highest prediction among all correct answers. 
(b) Visual verification utilizing answer semantics.%An example of VQA as VE.
}

  \label{fig:intro_image}
  \vspace{-0.2cm}
\end{figure}
%%%%%%%%%%%%%%%%%%%%%%%%%%%%%%%%%%%%%%%%%%%%%%%%%%%%%%%%%%%%%
As presented in Figure \ref{fig:intro_image}(a), quite a few correct answers usually occur at top $N$ candidates rather than top one. Meanwhile, if the top $N$ candidate answers are given, the image can further verify the visual presence/absence of concepts based on the combination of the question and the candidate answer. As shown in Figure \ref{fig:intro_image}(b), the question is about the color of the bat and two candidate answers are “yellow” and “black”. After checking the correctness of candidate answers, the wrong answer “yellow” which is contradicted with the image can be excluded and the correct answer “black” which is consistent with the image is confirmed. Nevertheless, this visual verification, which utilizes answer semantics to alleviate language priors, has not been fully investigated.

%Even if the model locates the correct region in the image, it 
%Due to the language priors problem, even if the model locates the correct region of the image, it may still give the top-1 prediction probability to those the high-frequency wrong answer top-1 prediction probability instead of the correct answer\citep{qiao2020rankvqa}. 
%Meanwhile, as shown in Figure\ref{fig:intro_image}(b), if the top $N$ candidate answers (including the correct answer “black” and wrong answer “yellow”) are given, the image can further verify the visual presence/absence of concepts based on the combination of the question and each answer. Then the answers “yellow” contradicted with the image can be excluded and the answer “black” which is consistent with the image is confirmed. This visual verification utilizing answer semantics has not been fully investigated. %encourages VQA models to %the question is about the color of the bat, and the high-frequency answer “yellow” is ranked first.
%the question is about the color of the bat, and the wrong answer “yellow” is ranked first. But after referring to the top $N$ candidate answers, those contradicted with the image can be excluded. Finally, the third candidate answer “black” which is consistent with the image is confirmed. 
%Nevertheless, the semantics of candidate answers have not been made good use of, and this visual verification, which utilizes answer semantics to %encourages VQA models to 
%alleviate the influence of data bias, has not been fully investigated.

%To address the above problems, 
In this paper, we propose a select-and-rerank (SAR) progressive framework based on Visual Entailment. The intuition behind the proposed framework comes from two observations. First, after excluding the answers unrelated to the question and image, the prediction space is shrunken %. For instance, if a question begun with “what color”, the candidate answers are almost categories related to colors. Thus,
and we can obtain a small number of candidate answers. 
Second, on the condition that a question and one of its candidate answer is bridged into a complete statement, the authenticity of this statement can be inferred by the content of the image. Therefore, after selecting several possible answers as candidates, %As a rule, a picture speaks a thousand words, thus 
we can utilize the visual entailment, consisting of image-text pairs, to verify whether the image semantically entails the synthetic statement. Based on the entailment degree, we can further rerank candidate answers and give the model another chance to find the right answer. %What's more, we develop an answer re-ranking module based on visual entailment to give the model another chance to find the right answer. %we rerank candidate answers by this visual entailment task and enable the model to find the right answer which is entailed by the image contents.
%The answer re-ranking based on visual entailment can gives another chance to find the right answer which could be entailed by the contents of image. %apart from all of the impossible, there is a truth left. Since the answers inconsistent with the image can be filtered out, the performance of overcoming language priors can be effectively improved.
To summarize, our contributions are as follows:

1. We propose a select-and-rerank progressive framework to tackle the language priors problem, %This framework makes full use of the interactive information of image, question and candidate answers. 
and empirically investigate a range of design choices for each module of this framework.
In addition, it is a generic framework, which can be easily combined with the existing VQA models and further boost their abilities.

2. We highlight the verification process between text and image, and formulate the VQA task as a visual entailment problem. This process makes full use of the interactive information of image, question and candidate answers.  %Through verifying whether the image semantically entails the text, the wrongly predicted answer can be eliminated and the bias problem can be alleviated.

3. %Extensive experiments are conducted on the widely used benchmark VQA-CP v2.
Experimental results demonstrate that our framework establishes a new state-of-the-art accuracy of $66.73\%$, outperforming the existing methods by a large margin. 

%%%%%%%%%%%%%%%%%%%%%%%%%%%%%%%%%%%%%%%%%%%%%%%%%%%%%%
\begin{figure*}
\resizebox{1.0\linewidth}{!}{
  \centering
  \includegraphics[width=1.0 \linewidth]{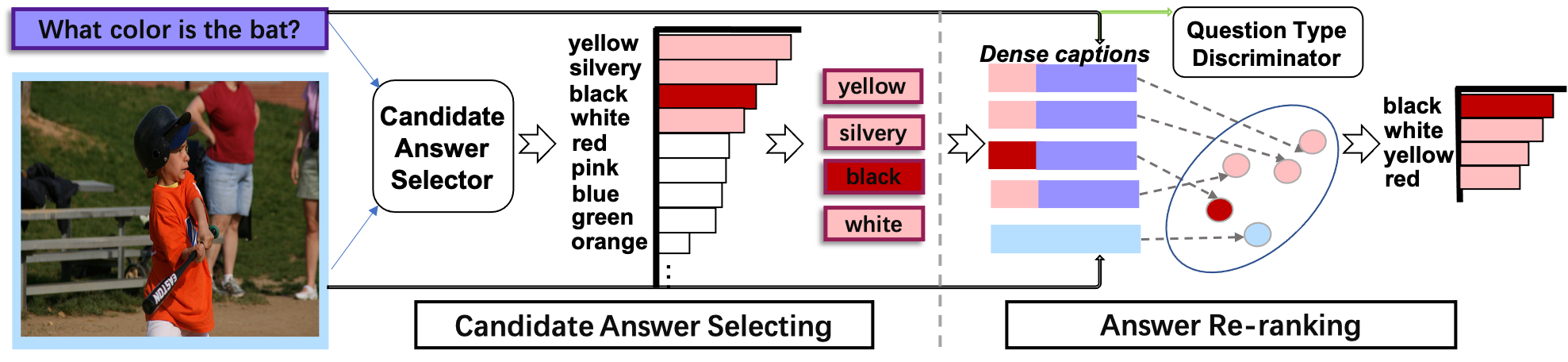}
  }
\caption{ Overview of the progressive framework SAR.  %Question Type Discriminator is used only in testing. %We combine the question with each candidate answer (selected by Candidate Answer Selector) as candidate dense captions. Then we match the visual representations of the image with the contextual representations of candidate dense captions to re-rank candidate answers. Intuitively, the more the image semantically entails the caption, the semantically closer the caption should be to the image. The candidate caption corresponding to the correct answer should be the closest one and we choose that answer as our final output.
}
\vspace{-0.1cm}
  \label{fig:model}
\end{figure*}
%%%%%%%%%%%%%%%%%%%%%%%%%%%%%%%%%%%%%%%%%%%%%%%%%%%%%%

\section{Related Work}
\paragraph{Language-Priors Methods}
%A few researchers \citep{jabri2016revisiting,agrawal2016analyzing,zhang2016yin,goyal2017making} have pointed out that most of current VQA models tend to answer questions by learning the superficial correlations in the data, without truly understanding the image contents. 
To address the language prior problem of VQA models, a lot of approaches have been proposed, which can be roughly categorized into two lines: %three kinds of solutions:
(1) \emph{Designing Specific Debiasing Models to Reduce Biases.} %This line aims to reduce biases by devising specific debiasing models, of which the most effective ones for VQA 
Most works of this line are ensemble-based methods \citep{ramakrishnan2018overcoming,grand2019adversarial,belinkov2019don,cadene2019rubi,clark2019don,mahabadi2019simple}, among these, LMH\citep{clark2019don} reduces all biases between question-answer pairs by penalizing the samples that can be answered without utilizing image content. %by penalizing the samples that can be answered without utilizing image content, LMH\citep{clark2019don} trains a robust model as a part of an ensemble with a naive model which is trained exclusively based on dataset biases, and thus learns an alternative strategy to reduce all biases between question-answer pairs. 
%2) Visual Supervision to Reduce Biases. These methods believe that the data biases problem is due to the model fails to utilize the correct visual regions to answer questions. Therefore, with the guidance of external visual supervision, they force the model to attend to the important visual regions and then predict answers. However, \citet{shrestha2020negative} has proved that these methods like SCR\citep{wu2019self} and HINT\citep{selvaraju2019taking} could reduce biases mainly because they has a regularization effect preventing the model from overfitting to linguistic priors. 
(2) \emph{Data Augmentation to Reduce Biases.} %The most straight-forward approach is data augmentation, in which more balanced datasets are carefully constructed to overcome language priors.
The main idea of such works  \citep{zhang2016yin,goyal2017making,agrawal2018don} is to carefully construct more balanced datasets to overcome priors. %For instance, \citet{zhang2016yin} and \citet{goyal2017making} improve the popular VQA dataset by collecting complementary images with opposite answers for each question and thus make it more balanced. %However, these “balanced” datasets can not thoroughly solve biases problem because the model can still leverage statistical biases from questions\citep{agrawal2018don}.
For example, the recent method SSL\citep{zhu2020overcoming} %is another typical data augmentation method, which alleviates biases by introducing 
first automatically generates a set of balanced question-image pairs, then introduces an auxiliary self-supervised task to use the balanced data. % without extra human annotations. 
CSS\citep{chen2020counterfactual} balances the data by adding more complementary samples which are generated by masking objects in the image or some keywords in the question. Based on CSS, CL\citep{liang2020learning} forces the model to utilize the relationship between complementary samples and original samples. %forces model to learn and full utilize the relationship between complementary samples and original samples.
Unlike SSL and CSS which do not use any extra manual annotations, MUTANT\citep{gokhale2020mutant} locates critical objects in the image and critical words in the question by utilizing the extra object-name labels, which directly helps the model to ground the textual concepts in the image.
However, above methods only explore the interaction between the image and the question, ignoring the semantics of candidate answers. In this paper, we propose a progressive VQA framework SAR which achieves better interaction among the question, the image and the answer.

\paragraph{Answer Re-ranking} Although
Answer Re-ranking is still in the infancy in VQA task, it has been widely studied for QA tasks like open-domain question answering, in which models need to answer questions based on a broad range of open-domains knowledge sources. 
%It takes a two-stage process for the open-domain question answering
Recent works \citep{wang2018joint,wang2018evidence,kratzwald2019rankqa} address this task in a two-stage manner: extract candidates from all passages, then focus on these candidate answers and rerank them to get a final answer. RankVQA\citep{qiao2020rankvqa} introduces Answer Re-ranking method to VQA task. They design an auxiliary task which reranks candidate answers according to their matching degrees with the input image and off-line generated image captions. %As beforementioned, \citep{qiao2020rankvqa} apply Answer Re-ranking to the VQA task. %Specifically, they use the relevance to visual content of the input image and the image captions to respectively rank all candidate answers. But it is worth noting that 
However, RankVQA still predicts the final answer from the huge prediction space rather than selected candidate answers. 

\section{Method}

Figure \ref{fig:model} shows an overview of the proposed select-and-rerank (SAR) framework, which consists of a Candidate Answer Selecting module and an Answer Re-ranking module. In the Candidate Answer Selecting module, given an image and a question, we first use a current VQA model to get a candidate answer set consisting of top $N$ answers. In this module, the answers irrelevant to the question can be filtered out. Next, we  formulate the VQA as a VE task in the Answer Re-ranking module, where  %we combine each candidate answer with the question, and refer the combination of the answer and the question as dense caption. In VE, 
the image is premise and the synthetic dense caption\citep{densecap} (combination of the answer and the question ) is hypothesis. %The image-caption pairs are then feed into the VE system based on the cross-domain pre-trained model LXMERT. 
We use  the cross-domain pre-trained model LXMERT\citep{tan2019lxmert} as VE scorer to %The VE system computes 
compute the entailment score of each image-caption pair, and thus the answer corresponding to the dense caption with the highest score is our final prediction. %The following sections will detail our proposed framework.

\subsection{Candidate Answer Selecting}

The Candidate Answer Selector (CAS) selects several answers from all possible answers as candidates and thus shrinks the huge prediction space. %The VQA model applied as CAS is a free choice in our framework. %and we directly use the existing VQA model. 
Given a VQA dataset $D=\{I_i,Q_i\}{_{i=1}^M}$ with $M$ samples, where $I_i \in I$, $Q_i \in Q$ are the image and question of the $i_{th}$ sample and $A$ is the whole prediction space consisting of thousands of answer categories. Essentially, the VQA model applied as CAS is a $|A|$-class classifier, and is a free choice in our framework. Given an image $I_i$ and a question $Q_i$, CAS first gives the regression scores over all optional answers: $P(A|Q_i,I_i)$. %Different from the routine of a traditional VQA model which directly outputs the answer with the highest score, CAS choose $N$ answers $A_i^*$ with top $N$ scores as candidates. 
%Then CAS chooses N answers $A_i^*$ with top $N$ scores as candidates, which is different from the routine of a VQA model which directly outputs the answer with the highest score.
%It is concluded as follows: 
Then CAS chooses $N$ answers $A_i^*$ with top $N$ scores as candidates, which is concluded as follows:
\begin{equation}
A_i^* = top N(argsort( P(A|Q_i,I_i)))
\end{equation}
%where $N$ (hyper-parameter) represents the number of candidate answers to be selected.  %And thus we get a set of N candidate answers $A_i^* = [A_i^1,A_i^2 , ..., A_i^N ]$  where we take each candidate answer $A_i^n$  as a word sequence.
$N$ (hyper-parameter) candidate answers $A_i^*= [A_i^1,A_i^2 , ..., A_i^N ]$ are selected for each $(I_i,Q_i)$ pair by CAS, forming a dataset $D^{'} = \{I_i, Q_i,A_i^n\}{_{i=1,n=1}^{M\ \ ,N}}$ with $M*N$ instances, where $A_i^n \in A_i^*$,  for the next Answer Re-ranking module. 
In this paper, we mainly use SSL as our CAS. We also conduct experiments to analyze the impact of different CAS and different $N$.
%In this paper, we mainly use SSL as our CAS and we also analyze the impact of different CAS and different $N$ with experiments respectively.
\subsection{Answer Re-ranking}
\subsubsection{Visual Entailment}
Visual Entailment (VE) task is proposed by \citet{xie2019visual}, where the premise is a real-world image, denoted by $P_{image}$, and the hypothesis is a text, denoted by $H_{text}$. Given a sample of ($P_{image}$, $H_{text}$), the goal of VE task is to determine whether the $H_{text}$ can be concluded based on the information of $P_{image}$. According to following protocols, the label of the sample is assigned to
(1)	\emph{Entailment}, if there is enough evidence in $P_{image}$ to conclude $H_{text}$ is true.
(2)	\emph{Contradiction}, if there is enough evidence in $P_{image}$ to conclude $H_{text}$ is false.
(3)	\emph{Neutral}, if there is no sufficient evidence in $P_{image}$ to give a conclusion about $H_{text}$.
%Recently, the single-stream pre-trained model UNITER \citep{chen2020uniter} achieves extraordinary performance boost on this task, which demonstrates the outstanding capability of cross-modal pre-trained models. Differently, in this paper, we adapt the two-stream model LXMERT\citep{tan2019lxmert} to respectively encode text and image , which aims to get a joint representation of question and answer before the cross-modal interaction.
\subsubsection{VQA As Visual Entailment}
%Given a question $Q_i$ and its candidate answers $A_i^*$, 
A question $Q_i$ and each of its candidate answers $A_i^*$ can be bridged into a complete statement, and then the image could verify the authenticity of each statement. More specifically, the visual presence of concepts (e.g. “black bat”/“yellow bat”) based on the combination of the question and the correct/wrong candidate answer can be entailed/contradicted by the content of the image. %For example, question “What color is the bat?” and answer “Black” could give a concept “black bat”. And we can determine that the answer is correct/wrong if the image entails/contradicts this concept. 
In this way, we achieve better interaction among question, image and answer. 

Therefore, we formulate VQA as a VE problem, in which the image $I_i$ is premise, and the synthetic statement of an answer $A_i^n$ in $A_i^*$ and question $Q_i$, represented as ($Q_i$,$A_i^n$), is hypothesis. For an image, synthetic statements of different questions describe different regions of the same image. Following \citet{densecap}, we also refer to the synthetic statement as “dense caption”. %In the VQA datasets, several questions may be answered based on different regions of same imgae, so each image corresponds to multiple dense captions.
We use $A_i^+$  to represent the $A_i^n$ if $A_i^n$ is the correct answer of $Q_i$, use $A_i^-$ otherwise. There is enough evidence in $I_i$ to prove ($Q_i$,$A_i^+$) is true, i.e. the visual linguistic semantically entails ($Q_i$,$A_i^+$). And there is enough evidence in $I_i$ to prove ($Q_i,A_i^-$) is false, i.e. the visual linguistic semantically contradicts ($Q_i,A_i^-$). Note that, there is no \emph{Neutral} label in our VE task and we only have two labels: \emph{Entailment} and \emph{Contradiction}. 

\subsubsection{Re-Ranking based on VE}
We re-rank dense captions by contrastive learning, that is, ($Q_i$,$A_i^+$) should be more semantically similar to $I_i$ than ($Q_i$,$A_i^-$). The right part of Figure \ref{fig:model} illustrates this idea.
%According to the idea of contrastive learning, when we match the visual representations of the image with the contextual representations of the candidate dense captions, % to rerank answers. 
%intuitively, the more the image $I_i$ semantically entails ($Q_i$,$A_i^n$), the semantically closer ($Q_i$,$A_i^n$) should be to $I_i$, while ($Q_i$,$A_i^+$) should be the closest one. 
The more semantically similar $I_i$ to ($Q_i$,$A_i^n$), the deeper the visual entailment degree is.
We score the visual entailment degree of $I_i$ to each ($Q_i$,$A_i^n$) $\in$ ($Q_i$,$A_i^*$)  and rerank the candidate answers $A_i^*$ by this score. The ranking-first answer is our final output. 
%According to the idea of contrastive learning, the extent to which the image $I_i$ entails ($Q_i$,$A_i^+$) is definitely much larger than the one to which $I_i$ entails ($Q_i$,$A_i^-$). Therefore, we use the VE extent to re-rank the candidate answers $A_i^*$ and finally choose a best answer. 
\paragraph{Question-Answer Combination Strategy}
The answer information makes sense only when combine it with the question. We encode the combination of question and answer text to obtain the joint concept.
%To accurately obtain the concept which is jointly expressed by the question and the answer, , which aimed at promoting better interaction between question and answer. 

We design three question-answer combination strategies: \textbf{R}, \textbf{C} and \textbf{R}$\rightarrow$\textbf{C} to combine question and answer into synthetic dense caption $C_i$:

\textbf{R}: \emph{Replace question category prefix with answer}. The prefix of each question is the question category such as “are there”, “what color”, etc. For instance, given a question “How many flowers in the vase?”, its answer “8” and its question category “how many”, the resulting dense caption is “8 flowers in the vase”. Similarly, “No a crosswalk” is the result of question “ Is this a crosswalk?” and answer “No”. We build a dictionary of all question categories of the train set, then we adopt a Forward Maximum Matching algorithm to determine the question category for every test sample.

\textbf{C}: \emph{Concatenate question and answer directly.}  For two cases above, the resulting dense captions are “8 How many flowers in the vase?” and “No Is this a crosswalk?”. The resulting dense captions after concatenation are actually rhetorical questions. We deliberately add answer text to the front of question text in order to avoid the answer being deleted when trimming dense captions to the same length.

\textbf{R}$\rightarrow$\textbf{C}: We first use strategy R at training, which is aimed at preventing the model from excessively focusing on the co-occurrence relation between question category and answer, and then use strategy C at testing to introduce more information for inference.

Adopting any strategy above, we combine $Q_i$ and each answer in $A_i^*$ to derive the dense captions $C_i^*$ . And thus we have a dataset $D^{''} = \{Ii, C_i^n\}{_{i=1,n=1}^{M\ \ ,N}}  $with $M*N$ instances for VE task.
\paragraph{VE Scorer}
We use the pre-trained model LXMERT to score the visual entailment degree of ($I_i$, $C_i^n$). LXMERT separately encodes image and caption text in two streams. Next, the separate streams interact through co-attentional transformer layers. In the textual stream, the dense caption is encoded into a high-level concept. %(e.g. 8 flowers, no crosswalk). 
Then the visual representations from visual stream can verify the visual presence/absence of the high-level concept.

We represent the VE score for the $i_{th}$ image and its $n_{th}$ candidate caption as: $sigmoid(Trm(I_i,C_i^n ))$, where $Trm ()$ is the 1-demensional output from the dense layers following LXMERT, $\delta$() denotes the sigmoid function. The larger score represents higher entailment degree.
We optimize parameters by minimizing the multi-label soft loss: %which is widely used in the existing VQA models:
\begin{equation}
\begin{aligned}
L_{VE} =&\frac{-1}{M*N}\sum_{i=1}^M\sum_{n=1}^N[t_i^nlog(\delta (Trm(I_i,C_i^n)))\\&+(1-t_i^n)log(1-\delta(Trm(I_i,C_i^n)))]
\end{aligned}
\end{equation}
where $t_i^n$ is the soft target score of the $n_{th}$ answer.% for the $i_{th}$ instance. 
\paragraph{Combination with Language-Priors Method}
After Candidate Answer Selecting, 
the amount of candidate answers decreases from all possible answers to top $N$. %This does not change the distribution of correct answers by question category because almost all correct answers are recalled by CAS during training. As a result, the most frequent answer for each question category is unchanged and 
Although some unrelated answers are filtered out, the dataset $D^{''}$ for VE system is still biased. Therefore, we can optionally apply existing language-priors methods to our framework for further reducing language priors. %even tho our framework itself possesses the capacity of alleviating language priors. 
Take the SSL as an example, we apply the loss function of its self-supervised task to our framework by adjusting the loss function to:
\begin{equation}
    L_{ssl}=\frac{\alpha}{M*N}\sum_{i=1}^M\sum_{n=1}^NP(I'_i,C_i^n)
\end{equation}
where $(I'_i,C_i^n)$ denotes the irrelevant image-caption pairs, $\alpha$ is a down-weighting coefficients. The probability $P(I'_i,C_i^n)$ could be considered as the confidence of $(I'_i,C_i^n)$ being a relevant pair. %This self-supervise task aims to prevent language priors from overly driving models by weakening the confidence of irrelevant pairs being relevant. The combination of language-bias method requires model to be training with a multi-task objective: visual entailment and language priors reduction. 
We can reformulate the overall loss function:
\begin{equation}
L = L_{VE} +  L_{ssl}
\end{equation}

\subsection{Inference Process}
%\subsubsection{Question Type Discriminator}
\paragraph{Question Type Discriminator}
%At the test stage, %an appropriate candidate answer number $N’$ matters a lot. On one hand, if CAS gives too many candidate answers(i.e. $N'$ is too large), it is more difficult for the model to choose the correct one among them. On the other hand, if $N'$ is too small, the correct answer with low regression scores is more likely to be omitted by the CAS. 
Intuitively, most “Yes/No” questions can be answered by the answer “Yes” or “No”. There is no need to provide too many candidate answers for “Yes/No” questions at the test stage. Therefore, we propose a Question Type Discriminator(QTD) to determine the question type and then correspondingly set different numbers of candidate answers, denoted as $N'$.  
%questions of different types require various candidate answers number, which matters a lot to performance. %Note that the question category (e.g. “what”,”how many”) is different from the question type (i.e. ”Yes\/No”, “Num” and “Other”). 
%The questions include 3 types: “Yes/No”, “Num” and “Other”. 
%We believe that “Yes/No” questions obviously demand fewer candidate answers than “Num” and “Other” questions. Therefore, for 
%For simplicity
Specifically, we roughly divided question types (including “Yes/No”, “Num” and “Other”) into yes/no and non-yes/no. A GRU binary classifier is trained with cross-entropy loss and evaluated with 5-fold cross-validation on the train split of each dataset. Then, the trained QTD model with an accuracy about 97\% is implemented as an off-line module during the test stage. %It is used to determine the question type and then correspondingly set different numbers of candidate answers, denoted as $N'$. %Following UpDn, QTD processes the question text with a GRU followed by a binary classifier. %This is a plug-in module at the test stage. 
We will further investigate the effect of  $N'$ on each question type in the next section.

%\subsubsection{Final Prediction}
\paragraph{Final Prediction}
In the inference phase, we search for the best dense caption $\hat{C_i}$ among all candidates $C_i^*$ for the $i_{th}$ image.
\begin{equation}
\hat{C_i}=\mathop{argmax}_{n \in N'}\delta(Trm(I_i,C_i^n))
\end{equation}
The answer $\hat{A_i}$ corresponding to $\hat{C_i}$ is the final prediction.

%%%%%%%%%%%%%%%%%%%%%%%%%%%%%%%%%%%%%%%%%%%%%%
\begin{table*}
\centering
\resizebox{0.95\linewidth}{!}{
\begin{tabular}{l|llll|llll|l}
\hline
%\textbf{SAR} & \textbf{UpDn-CAS} & \textbf{LMH-CAS}  & \textbf{SSL-CAS}\\

\hline 
\multirow{2}{*}{Model} &
\multicolumn{4}{|c|}{VQA-CP v2 test(\%)$\uparrow$}&\multicolumn{4}{|c|}{VQA-v2 val(\%)$\uparrow$}&
{GAP}\\\cline{2-9}
&ALL&	Yes$/$No	&Num&	Other&	All	&Yes$/$No	&Num&	Other&(\%)$\downarrow$ \\
\cline{1-10}
UpDN\citep{anderson2018bottom} & 39.74 & 42.27 & 11.93 & 46.05 & 63.48 & 81.18 & 42.14 & 55.66 & 23.74\\
Areg\citep{ramakrishnan2018overcoming} & 41.17 &	65.49 &	15.48	&35.48	&62.75&	79.84&	42.35&	55.16&	21.58 \\

RUBI\citep{cadene2019rubi} &47.11&	68.65&	20.28&	43.18&	61.16&	-&	-&	-&	14.05 \\
LMH\citep{clark2019don} & 52.45&	69.81&	44.46&	45.54&	61.64&	77.85&	40.03&	55.04&	9.19 \\\hline
RankVQA\citep{qiao2020rankvqa} & 43.05&	42.53&	13.91&	51.32&	65.42&	82.51&	\textbf{57.75}&	45.35&	22.37 \\ \hline
%HINT & 47.70	&70.04&	10.68&	46.31&	62.35&	80.49&	41.75&	54.01&	14.65 \\
%SCR&  48.47	&70.41&	10.42&	47.29&	62.30&	77.40&	40.90&	56.50&	13.83 \\ %\hline
LXMERT\citep{tan2019lxmert} & 46.23&	42.84&	18.91&	55.51&	\textbf{74.16}&	\textbf{89.31}&	56.85&	\textbf{65.14}&	27.93\\\hline
SSL\citep{zhu2020overcoming}& 57.59&	86.53&	29.87&	50.03&	63.73&	-&	-&	-&	6.14\\
CSS\citep{chen2020counterfactual} & 58.95&	84.37&	49.42&	48.21&	59.91&	73.25&	39.77&	55.11&	\textbf{0.96}\\
CL\citep{liang2020learning} &59.18 & \textbf{86.99} & 49.89 &47.16 & -&-&-&-&-\\
\hline
Top12-SAR(R$\rightarrow$C)\ \ \ (\textbf{Ours})	&64.55&	83.03&	50.05	&58.8&	70.41&	87.87&	54.34&	61.38&	5.86\\ 
Top20-SAR(R$\rightarrow$C)\ \ \ (\textbf{Ours})	&65.44	&83.13	&54.52&	\textbf{59.16}&	\underline{70.63}&	\underline{87.91}&	\underline{54.93}&	\underline{61.64}&	5.19\\ \hline

Top12-SAR+SSL(R$\rightarrow$C)\ (\textbf{Ours})&	64.29	&82.86&	51.98&	57.94&	69.84&	87.22&	54.41&	60.70&	5.55\\
Top20-SAR+SSL(R$\rightarrow$C)\ (\textbf{Ours})&	65.32&	83.41&	54.32&	58.85&	70.03&	87.47&	54.59&	60.85&	4.71\\ \hline

Top12-SAR+LMH(R)\ \ \ (\textbf{Ours})&	65.93&	85.38	&62.30&	56.73&	69.13	&87.61&	50.43&	60.03&	3.20\\
Top20-SAR+LMH(R)\ \ \ (\textbf{Ours})&	\textbf{66.73}	&\underline{86.00}&	\textbf{62.34}&	57.84&	69.22&	87.46&	51.20&	60.12&	\underline{2.49}\\

\hline
\end{tabular}}

\caption{\label{table:main}
Results on VQA-CP v2 test and VQA-v2 validation set. Overall best scores are bold, our best are underlined. The gap represents the accuracy difference between VQA v2 and VQA-CP v2. }
\vspace{-0.1cm}
\end{table*}

%%%%%%%%%%%%%%%%%%%%%%%%%%%%%%%%%%%%%%%%%%%%%%
\section{Experiments}
\subsection{Setting}
\paragraph{Datasets}
Our models are trained and evaluated on the VQA-CP v2\citep{agrawal2018don} dataset, which is well-crafted by re-organizing VQA v2\citep{goyal2017making} training and validation sets such that answers for each question category (65 categories according to the question prefix) have different distributions in the train and test sets. %Those frequent answers in the train set will rarely occur in the test set. 
Therefore, VQA-CP v2 is a natural choice for evaluating VQA model’s generalizability. The questions of VQA-CP v2 include 3 types: “Yes/No”, “Num” and “Other”. Note that the question type and question category (e.g.“what color”) are different. 
Besides, we also evaluate our models on the VQA v2 validation set for completeness, and compare the accuracy difference between two datasets with the standard VQA evaluation metric\citep{antol2015vqa}. 
\paragraph{Baselines}
We compare our method with the following baseline methods: UpDn\citep{anderson2018bottom}, AReg\citep{ramakrishnan2018overcoming}, RUBi\citep{cadene2019rubi}, LMH\citep{clark2019don}, %SCR\citep{wu2019self}, HINT\citep{selvaraju2019taking},
RankVQA\citep{qiao2020rankvqa}, SSL\citep{zhu2020overcoming}, CSS\citep{chen2020counterfactual}, CL\citep{liang2020learning} and LXMERT\citep{tan2019lxmert}. Most of them are designed for the language priors problem, while LXMERT represents the recent trend towards utilizing BERT-like pre-trained models\citep{li2019visualbert, chen2020uniter, li2020oscar} which have top performances on various downstream vision and language tasks (including VQA-v2). 
%are baselines representing data augmentation methods, as they achieve superior performance on language bias dataset. 
Note that MUTANT\citep{gokhale2020mutant}  uses the extra object-name label to ground the textual concepts in the image. For fair comparison, we do not compare with MUTANT.%choose SSL and CSS but MUTANT to be compared with our method.

\subsection{Implementation Details}
%The model applied as Candidate Answer Selector is a free choice in SAR framework.
%For the , one can choose any current VQA model.
In this paper, we mainly choose SSL as our CAS %To find out the performance limits of our methods under the choice of different CAS, we show the results of three models with UpDn(39.74), LMH(52.45), SSL(57.59), on the VQA-CP v2 test set. As beforementioned, most correct answers rank at front positons in all candidate answers. Top3 accuracy(acc) is about $70\%$ and top6 acc is $80\%$, which is far superior to the previous best. And thus, the performance drop resulted from failing to recall the correct answers is negligible. 
and set $N$=12 and $N$=20 for training. % during experiments. %All three CAS could achieve about $90\%$ top12 acc, $92\%$ top20 acc in training dataset and $86\%$ top12 acc , $88\%$ top20 acc in test. 
%The hyperparameter N at the training stage setting is further investigated in the next section. 
To extract image features, we follow previous work and use the pre-trained Faster R-CNN to encode each image as a set of fixed 36 objects with 2048-dimensional feature vectors. We use the tokenizer of LXMERT to segment each dense caption into words. All the questions are trimmed to the same length of 15 or 18, respectively for R or C question-answer combination strategy. In the Answer Re-ranking Module, we respectively incorporate two language-priors methods, SSL and LMH, into our proposed framework SAR, which is dubbed as SAR+SSL and SAR+LMH. Our models are trained on two TITAN RTX 24GB GPUs. We train SAR+SSL for 20 epochs with batch size of 32, SAR and SAR+LMH for 10 epochs with batch size of 64. For SAR+SSL, we follow the same setting as the original paper\citep{zhu2020overcoming}, except that we don’t need to pre-train the model with the VQA loss before fine-tuning it with the self-supervised loss. The Adam optimizer is adopted with the learning rate 1e–5.

For Question Type Discriminator, we use 300-dimensional Glove\citep{pennington2014glove} vectors to initialize word embeddings and feed them into a unidirectional GRU with 128 hidden units. When testing on the VAQ-CP v2, $N'$ ranges from 1-2 for yes/no questions and 5-15 for non-yes/no questions. As for VQA v2, $N'$ ranges from 1-2 for yes/no questions and 2-5 for non-yes/no questions.

\subsection{Results and Analysis}
\subsubsection{Main Results}
Performance on two benchmarks VQA-CP-v2 and VQA-v2 is shown in Table \ref{table:main}. We report the best results of SAR, SAR+SSL and SAR+LMH among 3 question-answer combination strategies respectively. “TopN-” represents that $N$ candidate answers (selected by CAS) feed into the Answer Re-ranking Module for training. Our approach is evaluated with two settings of $N$ (12 and 20). 

From the results on VQA-CP v2 shown in Table \ref{table:main}, we can observe that: (1) Top20-SAR+LMH establishes a new state-of-the-art accuracy of $66.73\%$ on VQA-CP v2, beating the previous best-performing method CL by $7.55\%$. Even without combining language-priors methods in Answer Re-ranking module, our model Top20-SAR  outperforms CL by $6.26\%$. These show the outstanding effectiveness of our proposed SAR framework. 
%Our method Top20-SAR+LMH establishes a new state-of-the-art accuracy on VQA-CP v2, beating the previous best-performing method CSS by $7.78\%$. Even without combining language-priors methods, our model Top20-SAR significantly outperforms CSS by $6.49\%$. %Two language-priors methods (SSL and LMH) both gain remarkable improvements after being incorporated into our framework. 
(2) SAR+SSL and SAR+LMH achieve much better performance than SSL and LMH, which demonstrates that SAR is compatible with current language-priors methods and could realize their full potential. 
%the great compatibility of SAR with current language-priors methods and also shows that SAR could realize full potential of these methods. 
%improve the performance of SSL and LMH %by $5.28\%$, $13.48\%$ with Top12 and $6.35\%$,$13.48\%$ with Top20. 
%This demonstrates the great compatibility of our framework with current language-priors methods, and our framework can achieve their full potential for overcoming language priors.  %(3) Our method boosts the original SOTA, CSS, by $6.49\%$ even without combining language-priors methods. 
(3) Compared with another reranking-based model RankVQA, our method elevates the performance by a large margin of $23.68\%$. This shows the superiority of our proposed progressive select-and-rerank framework over RankVQA which only uses the answer reranking as an auxiliary task. % to update model parameters. 
%This is because their work only uses the reranking as an auxiliary task to update model parameters, but our method carefully checks the authenticity of candidate answers during reranking process and outputs the ranking-first answer as final prediction. 
%does not get the final answer though answer ranking, and is actually a multi-task learning model rather than a model in select-and-rerank manner. 
(4) Previous models did not generalize well on all question types. CL is the previous best on the “Yes/No”, “Num” questions and LXMERT on the “Other” questions.
%CL and LXMERT respectively achieve the best performance on the “Yes/No”, “Num” and “Other” question types. 
In comparison, our model not only rivals the previous best model on the “Yes/No” questions but also improves the best performance on the “Num” and “Other” questions by $12.45\%$ and $3.65\%$. %It is worth noting that language priors are more likely to exist in the simple questions like “Yes/No” questions. “Other” questions require common sense. And more difficult “Num” questions test model’s inference abilities like visual grounding. 
The remarkable performance on all question types demonstrates that our model makes a significant progress toward a truly comprehensive VQA model.

We also evaluate our method on the VQA v2 which is deemed to have strong language biases. As shown in Table \ref{table:main}, our method achieves the best accuracy of $70.63\%$ amongst baselines specially designed for overcoming language priors, and is the closest to the SOTA established by LXMERT which is trained explicitly for the biased data setting. For completeness, the performance gap between two datasets is also compared in Table \ref{table:main} with the protocol from \citet{chen2020counterfactual}. Compared with most previous models which suffer severe performance drops between VQA v2 and VQA-CP v2 (e.g., %$23.74\%$ in UpDn, and 
$27.93\%$ in LXMERT), the Top20-SAR+LMH significantly decreases the performance drop to $2.49\%$, which demonstrates the effectiveness of our framework to further overcome the language biases. Though CSS achieves a better performance gap, it sacrifices the performance on the VQA v2. %Though CSS achieves better performance gap, its absolute performance on VQA v2 is not competitive with other methods. 
Meanwhile, as $N$ rises from 12 to 20, our models achieve better accuracy on both datasets along with a smaller performance gap. This demonstrates that, unlike previous methods, our method can alleviate language priors while maintaining an excellent capability of answering questions. Nonetheless, we believe that, how to improve the model’s generality and further transform the trade-off between eliminating language priors and answering questions into win–win outcomes, is a promising research direction in the future.

%Overall, our proposed method has a prominent accuracy boost with a large margin on VQA-CP v2, while maintaining competitive performance on the standard VQA v2 dataset compared to the SOTA by LXMERT.

%%%%%%%%%%%%%%%%%%%%%%%%%%%%%%%%%%%%%%%%%%%%%%%%%%%%%%
\begin{figure}
\resizebox{0.95\linewidth}{!}{
  \centering
  \includegraphics[width=1.0 \linewidth]{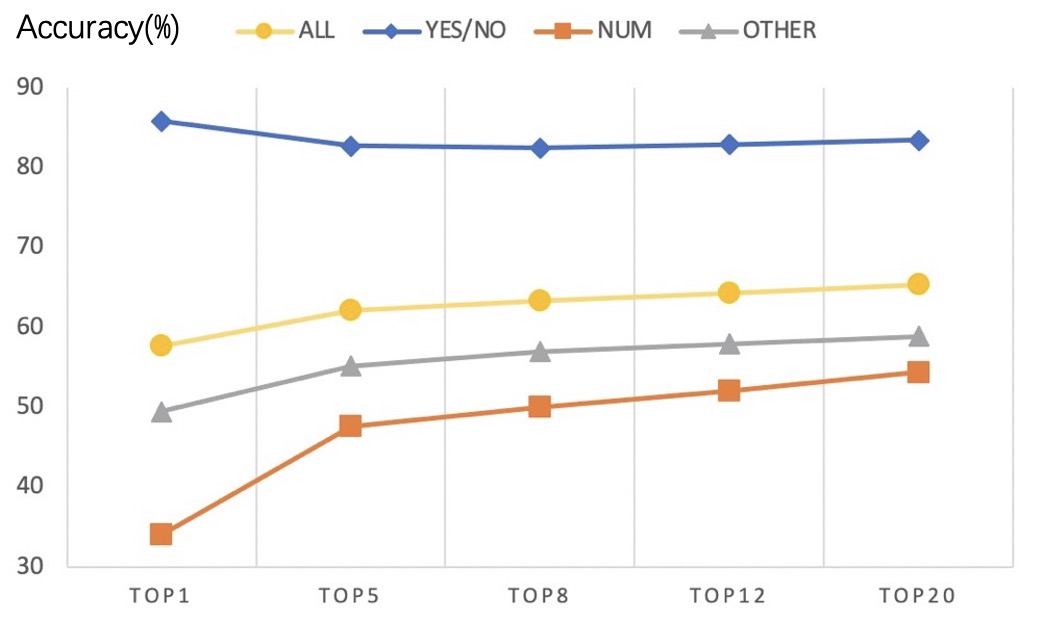}
  }

\caption{  \label{fig:Ntrain}
Results from model SAR+SSL(R$\rightarrow$C) in VQA-CP v2 with different $N$ during training.}
\vspace{-0.1cm}
\end{figure}
%%%%%%%%%%%%%%%%%%%%%%%%%%%%%%%%%%%%%%%%%%%%%%%%%%%%%%
\subsubsection{The Effect of $N$}

%To investigate the impact of the number of candidate answers $N$ during training, we conduct extensive experiments with different $N$ settings. Due to space limitations, we only analyze the case SAR+SSL using R$\rightarrow$C strategy, see Figure\ref{fig:Ntrain}. 
From %the results shown in 
Figure \ref{fig:Ntrain}, we can observe that the overall performance is getting better as $N$ increases. The performance improvement on the “Num” and “Other” questions is especially obvious, and there is a very slight drop on the “Yes/No” questions. %Based on the upward tendency shown in Figure\ref{fig:Ntrain}, w
We believe that SAR can further get better performance by properly increasing $N$. %using a larger $N$. 
%We believe that using an appropriate 
%one can further improve model performance by using a larger $N$. 
Due to the resource limitation, the largest $N$ we use is 20 in this paper. %and we mainly discuss the results based on $N$=12 in this paper.

%%%%%%%%%%%%%%%%%%%%%%%%%%%%%%%%%%%%%%%%%%%%%%
\begin{table}
\centering
\begin{tabular}{l|lll}
\hline
%\textbf{SAR} & \textbf{UpDn-CAS} & \textbf{LMH-CAS}  & \textbf{SSL-CAS}\\

Model$/$CAS & UpDn & LMH &SSL\\\hline
w/o SAR$^*$ & 41.04 & 53.03 & 57.66 \\\hline
SAR &61.71 & 61.65&64.55\\\hline
SAR+SSL & 63.52&	61.78&	64.29\\\hline
SAR+LMH & 64.98&	62.72&	65.14 \\\hline
\end{tabular}
   
\caption{ \label{table:CAS_diff}
Results %of SAR(R$\rightarrow$C), SAR+SSL (R$\rightarrow$C) and SAR+LMH(R$\rightarrow$C) 
based on different CAS in VQA-CP v2. We set N=12. $^*$ indicates the results come from our reimplementation using official released codes. }
\vspace{-0.1cm}
\end{table}
%%%%%%%%%%%%%%%%%%%%%%%%%%%%%%%%%%%%%%%%%%%%%%
\subsubsection{The Effect of Different CAS}
%We conduct experiments based on different Candidate Answer Selector: UpDn, LMH and SSL. 
To find out the potential performance limitation of CAS models, we show the accuracy of 3 CAS models on the VQA-CP v2 test set. As shown in Figure \ref{fig:intro_image} (a), the Top3 accuracy (acc) of 3 models is about $70\%$ and Top6 acc is $80\%$, which guarantees that sufficient correct answers are recalled by CAS. %And thus, the negative effect resulted from omitting very few correct answers is negligible.     
And thus, the performance limitation of CAS is negligible.  
%is far superior to the previous best. %And thus, the negative effect resulted from failing to recall enough %very few correct answers is negligible. 
%And thus, the negative effect resulted from omitting very few correct answers is negligible. 

%the negative effect resulted from failing to recall very few correct answers is negligible. 
%the performance drop resulted from failing to recall the correct answers is negligible.

We also conduct experiments to investigate the effect of different CAS on SAR. From the results shown in Table \ref{table:CAS_diff}, we can observe that: 
(1) Choosing a better VQA model as CAS does not guarantee a better performance, e.g. performance based on UpDn outperforms that based on LMH, but LMH is a better VQA model in overcoming language priors compared with UpDn. This is because a good Candidate Answer Selector has two requirements: (a) It should be able to recall more correct answers. (b) Under the scenario of language biases, wrong answers recalled by CAS at training time should have superficial correlations with the question as strong as possible. However, the ensemble methods, such as LMH, are trained to pay more attention to the samples which are not correctly answered by the question-only model. This seriously reduces the recall rate of those language-priors wrong answers, which leads to the training data for VE is too simple and thus hurts the model’s capability of reducing language priors. (2) If CAS is the general VQA model UpDn rather than LMH and SSL, the improvement brought from the combination with language-priors method in Answer Re-ranking module is more obvious. %Our framework could gain remarkable performance no matter with any CAS. 
(3) Even we choose the UpDn, a backbone model of most current works, as our CAS and \textbf{do not involve any language-priors methods}, SAR still achieves a much better accuracy than the previous SOTA model CL by $2.53\%$, which shows that our basic framework already possesses outstanding capability of reducing language priors. 
%As the performance of the model applied in CAS is stronger, the improvement of the combination of language-priors method in Answer Re-ranking module is less. 
%This is because, in the testing stage, CAS has eliminated some biased answers from the candidate answers, which makes the improvement room of language bias method decrease.
%%%%%%%%%%%%%%%%%%%%%%%%%%%%%%%%%%%%%%%%%%%%%%
\begin{table}
\centering
\resizebox{1\linewidth}{!}{
\begin{tabular}{l|l|lll}
\hline
%\textbf{SAR} & \textbf{UpDn-CAS} & \textbf{LMH-CAS}  & \textbf{SSL-CAS}\\

Top $N$ &Model & R & C &R$\rightarrow$C\\\hline
\multirow{3}{*}{Top12} &SAR&59.51&60.24&64.55\\
&SAR+SSL&	62.12&	62.87&	64.29\\
&SAR+LMH&	65.93&	65.23&	65.14\\ \hline
\multirow{3}{*}{Top20} &SAR&	60.43&	61.86&	65.44\\
&SAR+SSL&	62.29&	63.94&	65.32\\
&SAR+LMH&	66.73&	65.19&	66.71\\
\hline
\end{tabular}}
   
\caption{ \label{tableQAcomb}
Results on the VQA-CP v2 test set based on different question-answer combination strategies: R, C and R$\rightarrow$C. The major difference between R and C is whether keeping question prefix which includes 65 categories. }
\vspace{-0.1cm}
\end{table}
%%%%%%%%%%%%%%%%%%%%%%%%%%%%%%%%%%%%%%%%%%%%%%
\subsubsection{The Effect of Question-Answer Combination Strategies}
%We conduct experiments based on three Q-A combination strategies: R, C and R$\rightarrow$C. The major difference between R and C is whether keeping question prefix which includes 65 categories. 
From the results shown in Table \ref{tableQAcomb}, we can observe that:
(1) From overall results, R$\rightarrow$C achieves or rivals the best performance on three models. On average, R$\rightarrow$C outperforms C by $2.02\%$ which demonstrates avoiding the co-occurrence of  question category and answer during training time could effectively alleviate language priors; R$\rightarrow$C outperforms R by $2.41\%$ which indicates that the information of question category is useful in inference. (2) On the SAR and SAR+SSL, C consistently outperforms R, but on the SAR+LMH, we see opposite results. This is probably because our method  and the balancing-data method SSL could learn the positive bias resulted from the superficial correlations between question category and answer, which is useful for generalization, but the ensemble-based method LMH will attenuate positive bias during de-biasing process.
(3) Even without language priors method, SAR with R$\rightarrow$C rivals or outperforms the SAR+SSL and SAR+LMH with R or C, which shows that R$\rightarrow$C strategy could help the model to alleviate language priors. As a result, compared with R or C, our framework with R$\rightarrow$C only gains a slight performance improvement after using the same language-priors methods. 
%Under R or C strategy, our framework gains obvious performance boost after adopting LMH or SSL. %, which demonstrates the great compatibility of our framework with current language-priors methods. 

%But under the R$\rightarrow$C, the performance improvement is smaller after using same language-priors methods, which indicates that R$\rightarrow$C could possesses certain capacity of alleviating language priors.

%%%%%%%%%%%%%%%%%%%%%%%%%%%%%%%%%%%%%%%%%%%%%%
\begin{table}
\centering
%\resizebox{\text{90mm}
%\setlength{\tabcolsep}{0.2mm} {

\setlength\tabcolsep{3pt}
%\resizebox{\textwidth}{!}{   
\resizebox{0.95\linewidth}{!}{
\begin{tabular}{l|llll}
\hline
%\textbf{} & \textbf{UpDn-CAS} & \textbf{LMH-CAS}  & \textbf{SSL-CAS}\\
Model&	All&	Yes$/$No	&Num	&Other\\\hline
LXM&	46.23&	42.84&	18.91&	55.51\\\hline
LXM+SSL&	53.09&	55.07&	29.60&	58.50\\\hline
CAS+LXM(R)&	55.58&	70.91&	29.14&	54.81\\\hline
CAS+LXM+SSL(R)&	59.41&	76.60&	40.81&	55.51\\\hline
CAS+LXM+QTD(R)&	59.51&	83.20&	29.17&	55.42\\\hline
%CAS+UpDn+SSL+QTD(R)&	52.01&	81.57&	17.63&	45.94\\\hline
CAS+LXM+SSL+QTD(R)&	\textbf{62.12}&	\textbf{85.14}&	\textbf{41.63}&	\textbf{55.68}\\\hline
%CAS+LXM+SSL+QTD(R$\rightarrow$C)&	\textbf{64.29}	&82.86&	\textbf{51.98}&	\textbf{57.94}\\\hline
%\hline
\end{tabular}}
    
\caption{\label{ablationtab}
Ablation study to investigate the effect of each component of Top12-SAR+SSL: Candidate Answer Selector (CAS), LXMERT (LXM), Question Type Discriminator (QTD) and SSL.
}
\vspace{-0.1cm}
\end{table}
%%%%%%%%%%%%%%%%%%%%%%%%%%%%%%%%%%%%%%%%%%%%%%

\subsubsection{Ablation Study}
%We conduct ablation experiments on the Top12-SAR+SSL model to analyse the impact of different components in our framework (see Table\ref{ablationtab}). SAR+SSL consists of four main components: CAS, LXMERT, the combination of SSL, and the QTD module in the testing phase. (i.e. CAS+LXM+SSL+QTD). 
“CAS+” represents we use the select-and-rerank framework. %a current VQA model to select topN answers as candidate answers, and then use VE to re-rank candidate answers to output final prediction.
%Without “CAS+” represents that the model directly outputs final answer from the huge prediction space, which is the routine of traditional VQA models. 
From Table \ref{ablationtab}, we can find that:
(1)	LXM+SSL represents directly applying SSL to LXMERT. Its poor performance shows that the major contribution of our framework does not come from the combination of the language-priors method SSL and pre-trained model LXMERT. 
(2)	Compared with LXM and LXM+SSL, CAS+LXM and CAS+LXM+SSL respectively gain prominent performance boost of $9.35\%$ and $6.32\%$, which demonstrates the importance and effectiveness of our proposed select-and-rerank procedure.
(3)	CAS+LXM+QTD(R) and CAS+LXM+SSL+QTD(R) respectively outperform CAS+LXM(R) and CAS+LXM+SSL(R) by $3.93\%$ and $2.71\%$, which shows the contribution of QTD module. This further demonstrates that choosing appropriate $N'$ for different question types is a useful step for model performance.
%During inference time, QTD determines the question type and correspondingly chooses appropriate $N'$, which is a useful step for model performance. %The impact of $N'$ on each type question at testing time is further explored in the next subsection.
%(4)	After replacing the pre-trained model LXMERT with UpDn, the performance drops a lot. This shows LXMERT is an important component of our framework. (5)
(4) CAS+LXM+SSL+QTD improves the performance of CAS+LXM+QTD by $2.61\%$, which shows that current language-priors methods fit our framework well and could further improve performance. % But it is important to note that even not using language-prior methods, our framework could still beat previous SOTA results, highlighting the strong capability to overcome language priors of our basic framework.

%%%%%%%%%%%%%%%%%%%%%%%%%%%%%%%%%%%%%%%%%%%%%%%%%%%%%%
\begin{figure}
\resizebox{1.0\linewidth}{!}{
  \centering
  \includegraphics[width=1.0 \linewidth]{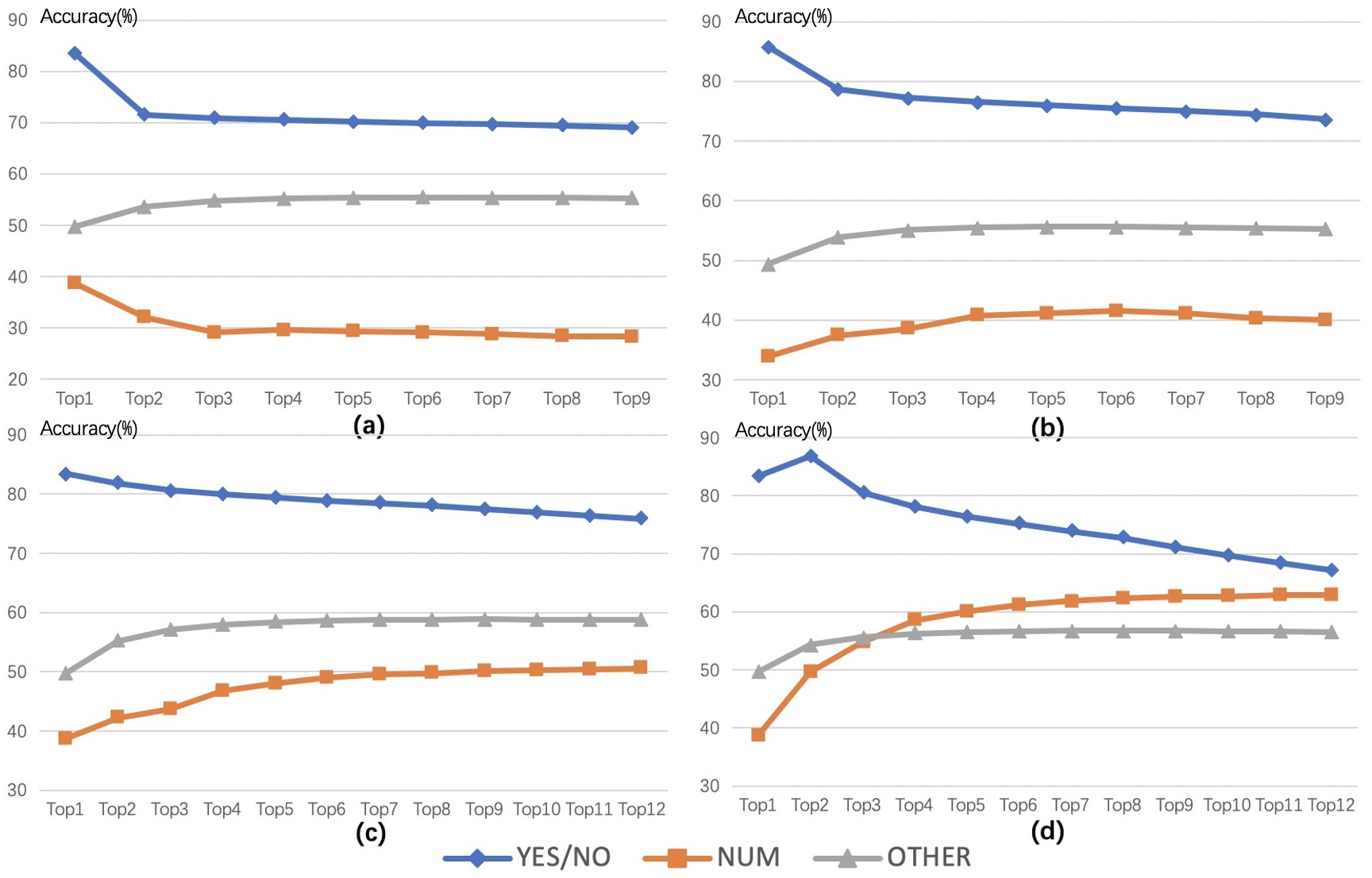}
  }
\caption{ Results from SAR(R), SAR+SSL(R), SAR(R$\rightarrow$C) and SAR+LMH(R) with different $N'$ during test.  To better investigate the impact of $N'$ on each question type, we report the results without Question Type Discriminator.}

  \label{fig:Ntest}
\vspace{-0.1cm}
\end{figure}
%%%%%%%%%%%%%%%%%%%%%%%%%%%%%%%%%%%%%%%%%%%%%%%%%%%%%%
%%%%%%%%%%%%%%%%%%%%%%%%%%%%%%%%%%%%%%%%%%%%%%%%%%%%%%
\begin{figure}
\resizebox{1.0\linewidth}{!}{
  \centering
  \includegraphics[width=1.0 \linewidth]{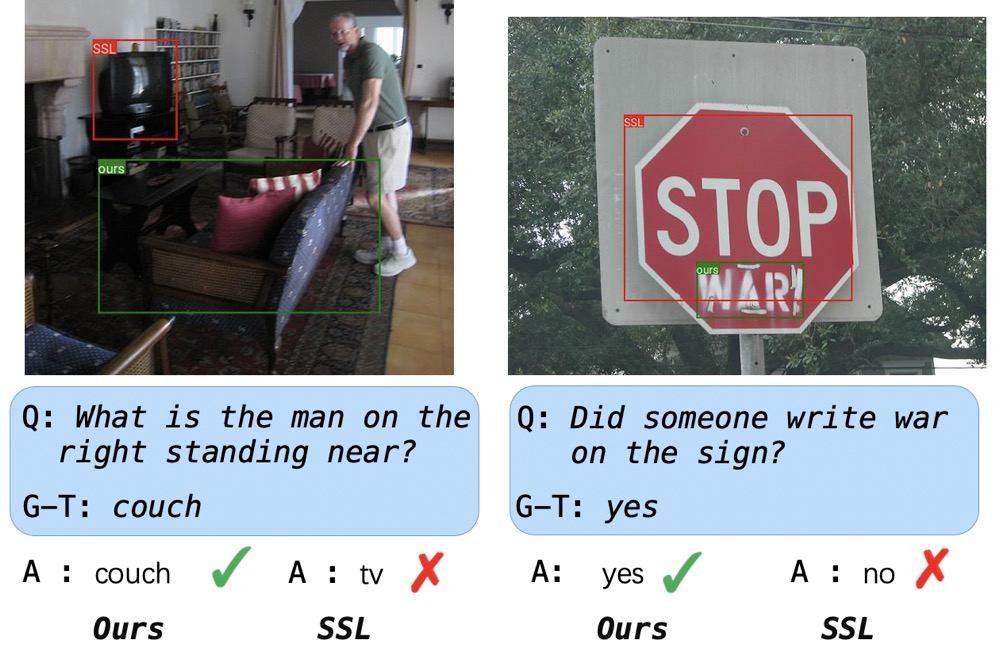}
  }
\caption{  Qualitative comparison between our Top20-SAR(R$\rightarrow$C) and the baseline SSL. The green/red bounding boxes indicate the most important regions resulting from ours/SSL. G-T is ground-truth.}

  \label{fig:visual}
\vspace{-0.1cm}
\end{figure}
%%%%%%%%%%%%%%%%%%%%%%%%%%%%%%%%%%%%%%%%%%%%%%%%%%%%%%

\subsubsection{The Effect of $N'$}
%Ablation study has already proved the important contribution of Question Type Discriminator. On the basic of QTD, we can correspondingly choose candidate answer number $N'$ for yes\/no question and non-yes\/no question respectively. 
%We conduct experiments to study the influence of different $N'$ on the three types of question. 
From Figure \ref{fig:Ntest}, we can find that: (1) The best $N'$ for yes/no questions is smaller than that for non-yes/no questions due to the nature of yes/no question. %For “Yes/No” questions, the best $N'$ is 1 or 2, as more candidate answers will only distract the model to make the correct choice. (2) %“Num” and “Other” questions are much harder, in which the best $N'$ is between 5 and 12. 
(2) As $N'$ increases, the accuracy of “Num” and “Other” questions rises first and then decreases. There is a trade-off behind this phenomenon: when $N'$ is too small, the correct answer may not be recalled by CAS; when $N'$ is too large, the distraction from wrong answers makes it more difficult for model to choose the correct answer.

%因此为不同类型选不同的答案是非常必要的. N对性能的影响是非常大的！！！
\subsubsection{Qualitative Examples}
 We qualitatively evaluate the effectiveness of our framework. As shown in Figure \ref{fig:visual}, compared with SSL, SAR performs better not only in question answering  but also in visual grounding. With the help of answer semantics, SAR can focus on the region relevant to the candidate answer and further use the region to verify its correctness. %For example, when answering the question “Is the person a man or a woman?”, our method can pay more attention to the person's hair, which might be an important visual clue to determine the gender of the person.
 
 %As shown in Figure\ref{fig:visual}, SAR performs well in answering correctly and visual grounding. This is because, under the help of answer semantics, it can focus on the region relevant to answers. % For example, %For example, when answering the question “Is this a professional game?”, our method can pay more attention to the characters on the man’s clothes, which might be an im- portant visual clue to judge whether the game is professional.

\section{Conclusion}
In this paper, we propose a select-and-rerank (SAR) progressive framework based on Visual Entailment. Specifically, we first select candidate answers to shrink the prediction space, then we rerank candidate answers by  a visual entailment task which verifies whether the image semantically entails the synthetic statement of the question and each candidate answer. Our framework can  make full use of the interactive information of image, question and candidate answers. In addition, it is a generic framework, which can be easily combined with the existing VQA models and further boost their abilities. 
We demonstrate advantages of our framework on the VQA-CP v2 dataset with extensive experiments and analyses. Our method establishes a new state-of-the-art accuracy of $66.73\%$ with an improvement of $7.55\%$ on the previous best.

\section*{Acknowledgments}
This work was supported by National Natural Science Foundation of China (No. 61976207, No. 61906187)

\bibliographystyle{acl_natbib}
\bibliography{anthology,acl2021}

%\appendix

\end{document}